\documentclass[conference]{IEEEtran}
	\IEEEoverridecommandlockouts
	\usepackage{cite}
	\usepackage{amsmath,amssymb,amsfonts}
	\usepackage{algorithmic}
	\usepackage{graphicx}
	\usepackage{textcomp}
	\usepackage{xcolor}
	\usepackage{pifont}
	\usepackage{stfloats}
	\def\BibTeX{{\rm B\kern-.05em{\sc i\kern-.025em b}\kern-.08em
		T\kern-.1667em\lower.7ex\hbox{E}\kern-.125emX}}
	\begin{document}

	\title{ISQuant: apply squant to the real deployment \\
	{\footnotesize \textsuperscript{*}Note: Sub-titles are not captured in Xplore and
	should not be used}
	\thanks{Identify applicable funding agency here. If none, delete this.}
	}

	\author{\IEEEauthorblockN{Dezan Zhao}}

	\maketitle

	\begin{abstract}
		The model quantization technique of deep neural networks has garnered significant attention and has proven to be highly useful in compressing model size, reducing computation costs, and accelerating inference. Many researchers employ fake quantization for analyzing or training the quantization process. However, fake quantization is not the final form for deployment, and there exists a gap between the academic setting and real-world deployment. Additionally, the inclusion of additional computation with scale and zero-point makes deployment a challenging task.

		In this study, we first analyze why the combination of quantization and dequantization is used to train the model and draw the conclusion that fake quantization research is reasonable due to the disappearance of weight gradients and the ability to approximate between fake and real quantization. Secondly, we propose ISQuant as a solution for deploying 8-bit models. ISQuant is fast and easy to use for most 8-bit models, requiring fewer parameters and less computation. ISQuant also inherits the advantages of SQuant, such as not requiring training data and being very fast at the first level of quantization. Finally We conduct some experiments and found the results is acceptable.our code is available at https://github.com/
	\end{abstract}

	\begin{IEEEkeywords}
	quantization
	\end{IEEEkeywords}

	\section{Introduction}
	Model compression\textsuperscript{\cite{survey}} now is nessesary for model acceleration and deployment. Althoght the computation ability and memory of hardware become bigger and stronger, The neural network become bigger and bigger. the deployment is still a problem. To get faster rate and less enermy need, Model compression is still needed. The model compression technique includes prune \textsuperscript{\cite{OBD}} CHEX \textsuperscript{\cite{CHEX}} Ps and Qs \textsuperscript{\cite{Ps_and_Qs}}  Once for All \textsuperscript{\cite{Once_for_All}} Any-precision \textsuperscript{\cite{Any_precision}}, quantization, distill, the light nerual network design, and nas. the quantization often is the Compatible technique for other compression technique. For example, we can prune and quantize the same time, etc. the deep compression \textsuperscript{\cite{Deep_compression}} do quantization after the prune, the Apq\textsuperscript{\cite{Apq}} Joint search for network architecture. After we do the distil we can quantize. What's more, we can do the distill to propose the accuray after the quantization.inference of neural networks. 
	To reduce the memory bandwidth and foot print, model size is always needed to compression. The reason why we can do model compression is that the models is usually over-parameterized. After the quantization, less FLOPS and less memory footprint is needed, and the inference's rate is faster.

	The Quantization has achieve great progress recent years. The bits become lower and accuracy become higher. From BWN \textsuperscript{\cite{Xnor-net}}, TWN \textsuperscript{\cite{TWN}}, Xnor-net \textsuperscript{\cite{Xnor-net}},  we must take the whole network and finetune, the AdaQuant \textsuperscript{\cite{AdaQuant}} ACIQ\textsuperscript{\cite{ACIQ}}, QDrop \textsuperscript{\cite{QDrop}} only use some train and not requires the label. What's more, train data is not nessesary For some work.  The work squant has pushes the accuracy and processing time of data free quantization to a new frontier. AdaRound \textsuperscript{\cite{AdaRound}} research found that the how round affect the accuracy.DFQ \textsuperscript{\cite{DFQ}} can do the quantization without the train data.
	SQuant \textsuperscript{\cite{DFQ}} is also without the train data and can calibrate faster.

	However, most work focus on the fake quantization accuracy, which is not the last form to deployment.\cite{MQBench} do the real deployment on hardware different platforms, found that there is a nonnegligible accuracy gap between the real deployment and academic setting. 

	a very hot directions is dummed as extreme bits quantization which can train the model into bnn, or bwn. 4 bits also can be see n good performance\textsuperscript{\cite{4bits}}\textsuperscript{\cite{tvm}}. But the performance may drop sharply. we can also use the fixed-precision which for different layer allocate different bits num. But many devices can only support 8bits acceleration. So we adpot 8bits as our objective bits num. The 8bits still is the common bits choice for application. Because the lower bits, the lower accuracy. What's more, many hardware device supports 8bits gemm acceleration. So We take the 8bits as our object bits.
	
	we can carry on 8bits quantization with  QAT, etc google's earlier work\textsuperscript{\cite{tensorflow_int8}}, F8Net\textsuperscript{\cite{F8Net}}, also can do it with PTQ, such as EasyQuant\textsuperscript{\cite{EasyQuant}}.
	
	because the privacy of data, some work need to do quantization without train data. The famous work is DFQ\textsuperscript{\cite{DFQ}}, Squant\cite{Squant} 
	
	Some works focus on the special nerual network architectures.do quantization with transformer\textsuperscript{\cite{transformer}}.
	
	The Hessian-based\textsuperscript{\cite{Hessian-based}} method is a promising direction, related work is Hawq\textsuperscript{\cite{Hawq}},Hawq-v3\textsuperscript{\cite{Hawq-v3}}, Squant\textsuperscript{\cite{Squant}}

	Here are our contribution below:
	\begin{itemize}
		\item we explain why use the fake quantization to analyse and train.
		\item we develop the squant to isquant, with less parameters and less computation cost.
		\item we conduct experiments to prove the ISQuant and cause negligible relative performance drop,  inherit the squant's fast quantization rate and no need for train data.
	\end{itemize}
	Deploment work is expected to get some inspiration from this work.

	\section{PRELIMINARIES}
		\subsection{notations}
		\textbf{number set}: a set of numbers which are the same type. The number set can be a tensor or a part of tensor. For example, we can think the whole Filters of a convolution is a number set, or we can think that every filter of convolution is a different number set. The function of the tensor can be ignored, the distribution characteristic of data is considered. 
		
		\textbf{signed number set}: the numbers in the number set is signed, some numbers are negative and some numbers are positive, the others are zero. For example, The weight tensor is signed number.
		
		\textbf{unsigned number set}: the numbers in the number set is all positive and zero. There is no negative number is in the set. For example, the activation tensor after the relu operation is a unsigned number set.

		\textbf{Transform Zone}: from a number set of one type to a number set of another type. We can classify the zone into there zones:the real zone, the quantized zone and the dequantized zone, respectively are float type, integer type and float type. We can call them the real number set, the quantized number set and the dequantized number set.
		
		we can describe the quantization as two transformation. 
		
		\textbf{quantize}: Transpose a number set from the real zone to the quantized zone. 
		
		\textbf{deqauntize}: Transpose a number set from the quantized zone to the dequantized zone.

		\textbf{quantization error} : the error between the real number set and the dequantized number set.
		
		\subsection{taxonomy}
		we can devide quantization technique into different categories according different norm. 
		\begin{itemize}
			\item \textbf{uniform quantization and non-uniform quantization}. 
			google's whitepaper\textsuperscript{\cite{whitepaper}}.thinking two number sets $A$ and $B$: we may set the range of A is $[a_{0}, a_{1}]$, and the range of B is $[b_{0}, b_{1}]$. The uniform quantization depends on special hardware. This paper only consider the uniform quantization.
			\item \textbf{symmetric quantization and asymmetric quantization}. 
			\item \textbf{static quantization and dynamic quantization}. For dynamic quantization, the zero point and the scale factor is computed for every input. The static quantization zero point and scale factor is computed once, then input that follow can share this zero point and scale factor. In this paper, we use the static quantization. The dynamic quantization theoretically can achieve higher accuraccy because the quantization parameter is computed for the specific data. But it will increase the computation cost.
			\item \textbf{weight quantization and activation quantization}. The quantized number set is weight tensor of conv or fc, called weight quantization. The quantized number set is activation tensor, we call activation quantization. If we only quantize the weight, called weight only quantization. To reduce static memory footprint of model size not during inference. ususally weight quantization and activation quantization is done at the same time.
			\item \textbf{post training quantization(PTQ) and quantization-aware training(QAT)}. PTQ means quantization during the training, it requires all the train dataset just like train the float model. PTQ means do quantization after the training, it does not need the train label but some train datasets.
			\item \textbf{per-channel quantization and per-tensor quantization}. The per-channel means every filter of one convolution has one scale coresponding to activation per-channel. per-tensor means the filtes, as a tensor, share a scale factor. 
			\item \textbf{extremely low bit and mixed-precision quantization}
			we call 1bits quantization and 2bits is extremely low bit quantization. The extremely low bit quantization causes the nonnegligible 
 			accuracy drop. Now 4bits can get a result which is almost equivalent to original accuracy by QAT methods, but the real deployment accuracy also drops according to \textsuperscript{\cite{MQBench}}.To get a trade-off performance between accuracy and bits number, mixed-precision quantization is applied. The mixed-precision decides bits of a layer according to its sensitivity to bits number. academic research is always make the first layer of model unquantized or allocate a big bits number, because generally speaking, the first layer is every sensitive to the bits number. Hawq-v2\textsuperscript{\cite{Hawq-v2}} shows that it is a good sensitivity metric to compute the average of all of the Hessian eigenvalues. 
		\end{itemize}
		
		we can also classify according to the theory. the DSQ classify the quantization solutions into four levels: 
		\begin{itemize}
			\item no data and no backpropogation
			\item requires data and no backpropagation
			\item requires data and requires backpropagation but just need fineting
			\item requires data and need backpropagation but needs to be trained from scratch
		\end{itemize}
		The first level, also need data, but the data is random data, like squant or generated from some rules.
		 The second level, often named Post-training quantization, named as PTQ, but some work, like Adaround\textsuperscript{\cite{AdaRound}}, QDrop\textsuperscript{\cite{QDrop}}, Brecq\textsuperscript{\cite{Brecq}}, still needs backprogation, but needs some train data, and use the original model's output as the label. So we describe the second levels as requires some train data and no train label.The three level, named as quantization-awre training, abbre as QAT,like Pact\textsuperscript{\cite{Pact}},LSQ\textsuperscript{\cite{LSQ}}, and LSQ+\textsuperscript{\cite{LSQ+}} and so on, has push the quantization training to new frontier.
		 so we depict the quantization level as the four categories more precisely. The proposed ISQuant belongs to the first level.
		 \begin{itemize}
			\item no train data and no train label.
			\item need some train data and no train label.
			\item need whole train data and train label but finetune.
			\item need whole train data and train label needs to be trained from scratch.
		\end{itemize}

		The paper F8Net\textsuperscript{\cite{F8Net}} classify the quantization into three categories: simulated quant, Inter-only quant and Fixed quant, this paper we concern the simulated quant, because the simulated quant is an intermidiate stage, with the convolution operation is integer, and can extend to other quant. We hope our work can give people some thought.

		\subsection{squant}
		squant's knowledge.
		squant adpot the Hessian-based approach which is one of most promising analysis methods to improve the performance of quantization.The optimization is 
		squant has three goals.firstly, make very weight element quantized, at the goal absolute quantization error smaller than 0.5, called EQ. This is the primitive nearest-round method.
secondly, make absolute sum of error(ASE) is smaller than 0.5, called KQ.
Thirdly, called CQ.
	\section{Methodology}
	as \cite{nvidia_empirical} depicted, the quantization can be classified into the affine quantization and scale quantization.
		\begin{gather}
			s = \frac{2^b - 1}{ \alpha  - \beta} \\
			z =  -round(\beta * s) - 2 ^{b -1}
		\end{gather}
		quantized zone's data range size is $2^b - 1$ for b bits quantization, and the real zone's data range size is $\alpha - \beta$. 
		Here s means the scale factor and the z means the zero-point, zero-point is the integer number in the quantized zone that the zero number in the real zone maps to.
		
		for the scale quantization
		\begin{equation}
			s = \frac{2^b - 1}{\alpha}
		\end{equation}
		for signed number set,
		\begin{equation}
		\begin{split}
			x_{q} &= quantize(x, b, s) \\
			      &= clip(round(s, x), -2^{b - 1} + 1, 2^{b - 1} - 1)
		\end{split}
		\end{equation}
		
		for unsigned number set,
		\begin{equation}
			\begin{split}
				x_{q} &= quantize(x, b, s) \\
				      &= clip(round(s, x), 0, 2^{b})	
			\end{split}
		\end{equation}
		the scale quantization is considered as the special situation where $\beta = 0$.

	the picture and be shown about the training process and the real:
	as \cite{MQBench} depicted, the QAT can be approximated as 
		\begin{equation}
			y_{ij} = \sum_{k = 1}^n \hat{w}_{ik} \hat{x}_{kj} = s_{w}s_{x} \sum_{k = 1}^n(\overline{w}_{ik}\overline{x}_{kj} - z_{w}\overline{x}_{kj} - z_{x}\overline{w}_{ik} + z_{w}z_{x})
		\end{equation}
		the computation is complex and destroy original convolution operation because the additional operation.
		consider the simple convolution example as Fig. 1.
		\begin{figure}[htbp]
			\centerline{\includegraphics[width=1\linewidth]{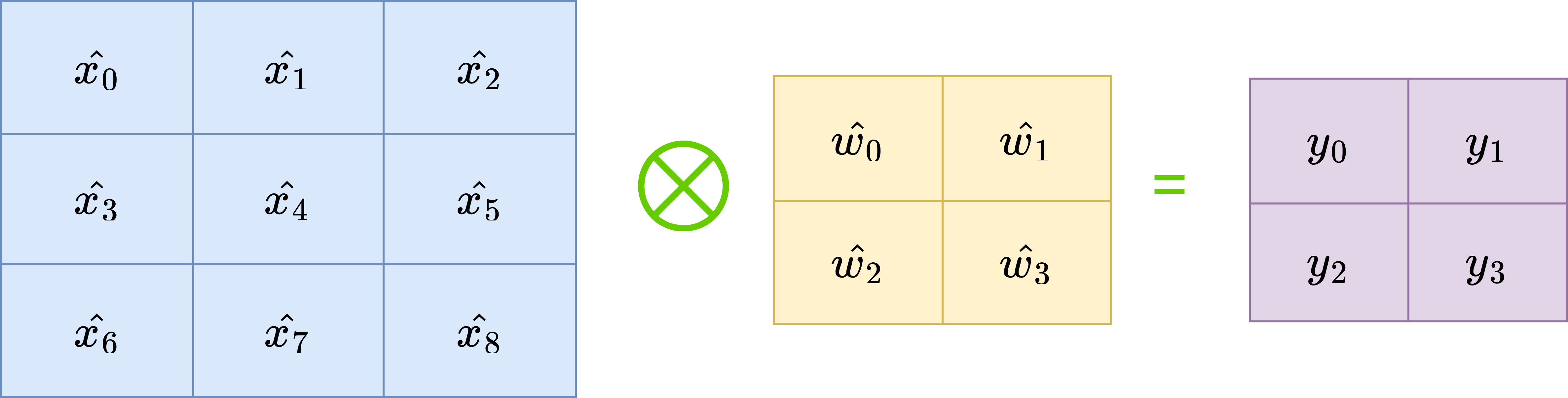}}
				\caption{Example of a Convolution.}
			\label{fig}
		\end{figure}
		take this for example.
		
		\begin{equation}
			\begin{split}
			 y_{0} &= \hat{x}_{0} * \hat{w}_{0} +  \hat{x}_{1} * \hat{w}_{1} + \hat{x}_{3} * \hat{w}_{2} +  \hat{x}_{4} * \hat{w}_{3} \\
				  &= ({x}_{0} - z_{x}) * s_{x} * ({w}_{0} - z_{w}) * s_{w} + \\
				 &\quad	(x_{1} - z_{x}) * s_{x} * ({w}_{1} - z_{w}) * s_{w} +  \\
				 &\quad	(x_{3} - z_{x}) * s_{x} * (w_{2} - z_{w}) * s_{w} + \\
				 &\quad	(x_{4} - z_{x}) * s_{x} * (w_{3} - z_{w}) * s_{w} \\
				  &= s_{x}s_{w}(p_{0} + p_{1} + p_{2} + p_{3})				  
			\end{split}
		\end{equation}
		where
		\begin{gather}
			p_{0} = x_{0}w_{0} + x_{1}w_{1} + x_{3}w_{2} + x_{4}w_{2} \\
			p_{1} = z_{w}(x_{0} + x_{1} + x_{3} + x_{4}) \\
			p_{2} = z_{x}(w_{0} + w_{1} + w_{2} + w_{3}) \\
			p_{3} = 4z_{x}z_{w}
		\end{gather}
		the item $p_{1}, p_{2}, p_{3}$ destroy the original convolution structure.

	it is hard to deploy because the zero-point, which incurs additional computation.the example is this. for the 3*3 input and 2*2 conv, and the output, the picture is like below. 
		\begin{equation}
		conv_Out = sdfdf
		\end{equation}
		\begin{figure}[htbp]
			\centerline{\includegraphics[width=0.8\linewidth]{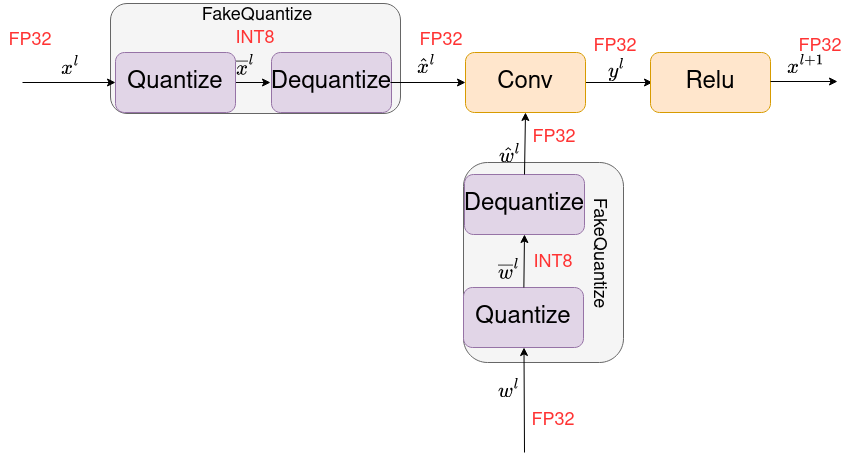}}
				\caption{Example of a figure caption.}
			\label{fig}
		\end{figure}
		
		\begin{figure}[htbp]
			\centerline{\includegraphics[width=1\linewidth]{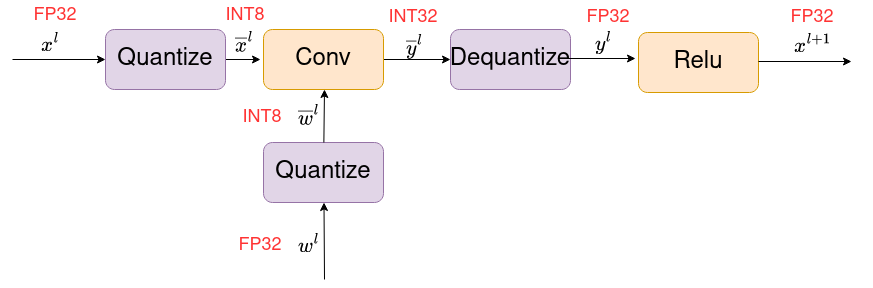}}
				\caption{Example of a figure caption.}
			\label{fig}
		\end{figure}
		
		why we consider the fake quantization to analyse the quantization problem. we can explain like this.
		\begin{itemize}
			\item approximate is good, we can use the tailor to.
			\item can train. if we take the real deployment runtime of quantization, when the loss backward,
			\begin{equation}
				y^l = \frac{\overline{y}_{l}}{s_{w} * s_{l}} 
			\end{equation}
			and 
			\begin{equation}
				\frac{\partial loss}{\partial \overline{y}_{l}} 
				 = \frac{\partial loss}{\partial y_{l}} * \frac{\partial   y_{l}}{\partial \overline{y}_{l}}
			\end{equation}
			then
			\begin{equation}
				\frac{\partial loss}{\partial \overline{y}_{l}} =\frac{\partial loss}{\partial y_{l}} * \frac{1}{s_{w} * s_{l}}
			\end{equation}
			$s_{w} * s_{l}$ is always a big number, and because the grad disappear, the $\frac{\partial loss}{\partial y_{l}}$ is ususally a small value, which leads to the $\frac{\partial loss}{\partial \overline{y}_{l}}$ tend to zero, the weight will not update. with the fake quantization We can use STE\textsuperscript{\cite{STE}} to do the backpropagation. 
		\end{itemize}
	however, the item with zero point is always hard to computer, so we relax the equation to as the zero-point to real 0, then the equation will become
		\begin{equation}
			y_{ij} = \sum_{k = 1}^n \hat{w}_{ik} \hat{x}_{kj} = s_{w}s_{x} \sum_{k = 1}^n(\overline{w}_{ik}\overline{x}_{kj})
		\end{equation}
		the equation is easier to do  and  can be used.
		
		we develop Squant as ISquant, which is faster to quantize and can do the quantization with free data.
		
		we use 8bit as the destination goal is because that it is the widely used bits. and lower bits may lead to big drop, which is sensitive to the symmetric and per-layer.
		
		our scheme can be describe like this:
		\begin{itemize}
			\item use symmetric not asymmetric. Zero-point is still set to zero.
			\item use per-tensor. per-tensor is less some parameters that per-channel, but reduce many computation.
			\item use 8bits.
			\item ues squant. based on squant enable us  to do quantization very fast and gain good performance without train dataset.
		\end{itemize}
		
		the advantage of our scheme is like this:
		\begin{itemize}
			\item easy to deploy, can use the original structure of conv
			\item less parameters and less computation. ISQuant change the perchannel scheme to per-tensor scheme, and change the asymmetric to symmetric form, reduce the paramzer
			\item fast, squant's advantage.
		\end{itemize}
		
		Folding Bn:
		\cite{MQBench}introduces four strategies for Batch Normlization Fold. Because our work does not need train, so we just fold the BN layers use the naive style.
		\begin{gather}
			w_{fold} = w \frac{\gamma}{\sqrt{\sigma^2 + \epsilon}} \\
			b_{fold} = \beta + (b - \mu)\frac{\gamma}{\sqrt{\sigma^2 + \epsilon}}
		\end{gather}
		$w_{fold}$ means the folded weight,$b_{fold}$means the folded bias, $w$means the original weight, $b$ means the original bias, $u$ is the running mean, $\sigma^2$ is variance, $gamma$ is the affine weight, $\beta$ is the affine bias, $\epsilon$ is a small value for the numerical stability. After the operation, BN layers is removed. We do the fuse operation before the quantization as a pre-process.
	\section{Results}
	\begin{table*}[htbp]
	\caption{8 bits quantization of different methods}
	\centering
	\begin{center}
	\begin{tabular}{|l|l|l|l|l|l|l|l|l|l|l|}
	\hline
			 &PerLayer&symmetric& FBN& Conv Int        & No BP & No SD & resnet18 & resnet50 & inceptionv3 & SqueezeNext \\ \hline
	Baseline &&&     &           &    &     & 71.47    & 77.74    & 78.81       & 69.38       \\ \hline
	DFQ      &\color{red}\ding{55}&\color{red}\ding{55}&\color{red}\ding{55}	 &        \color{red}\ding{55}      &  \color{green}\checkmark  &  \color{green}\checkmark   & 69.70    & 77.65    & -           & -           \\ \hline
	ZeroQ    &\color{red}\ding{55}&\color{red}\ding{55}&\color{red}\ding{55}	 &      \color{red}\ding{55}        & \color{red}\ding{55}   & \color{green}\checkmark  & 71.43    & 77.68    & 78.78       & 68.18       \\ \hline
	GDFQ     &\color{red}\ding{55}&\color{red}\ding{55}&\color{red}\ding{55}&      \color{red}\ding{55}        & \color{red}\ding{55}   & \color{red}\ding{55}                & 70.68    & 77.51    & 78.62       & 68.22       \\ \hline
	SQuant   &\color{red}\ding{55}&\color{red}\ding{55}&\color{red}\ding{55}	 &       \color{red}\ding{55}       & \color{green}\checkmark   &  \color{green}\checkmark   & 71.47    & 77.71    & 78.79       & 69.22       \\ \hline
	ISQuant  &\color{green}\checkmark&\color{green}\checkmark&\color{green}\checkmark& \color{green}\checkmark & \color{green}\checkmark   &  \color{green}\checkmark   & 71.046   & 77.140   & 78.688      & 68.342      \\ \hline
	\multicolumn{11}{l}{FBN mean fold batch norm, No BP means no backprogation, No SD means no synthetic data, Conv Int means convolution computation use integer.}
	\end{tabular}
	\end{center}
	\end{table*}
	We firstly study the accuracy of the different algorithms on the dataset imagenet on 8 bits number setting. From Table I, we can see that even our ISQuant satisfy all the six conditions, the relative error still drop negligibly. The resnet18 is on the $ >= 71\%$, resnet50 $> 77\%$, inceptionv3 $>=78\%$, and the SqueezeNext\textsuperscript{\cite{SqueezeNext}} is $>=68\%$. The SqueezeNext drop about $0.9\%$ compare the squant, that's because network architecture of SqueezeNext is the light network, which is more sensitive to the zero-point and per-layer settings. Note the ISQuant has fuse the batchnorm into convolution, and the convolution operate with the int8. we can see that although our ISQuant use convolution int, still get good performance.
	\begin{table*}[htbp]
	\caption{other bits performance of ISQuant}
	\begin{center}
	\begin{tabular}{|l|l|l|l|l|l|l|l|l|}
	\hline
			 & resnet18 & resnet34 & resnet50 & resnet101 &VGG19   & InceptionV3 &InceptionV4 & SqueezeNext \\ \hline
	Baseline & 71.47    & 75.178    & 77.74    & 78.102   &73.810  & 78.81      &79.392      & 69.38       \\ \hline
	8bit     & 71.046\color{red}(0.424) & 75.144\color{red}(0.034) & 77.140\color{red}(0.599)   & 77.974\color{red}(0.128)  &73.102\color{red}(0.707) & 78.688\color{red}(0.121)      & 79.120\color{red}(0.272)& 68.342\color{red}(1.038)      \\ \hline
	7bit     & 70.786\color{red}(0.684) & 74.904\color{red}(0.274) & 76.846\color{red}(0.893)   & 77.872\color{red}(0.23)   & 73.108\color{red}(0.702)    & 78.478\color{red}(0.332)      & 78.968\color{red}(0.423)& 67.390\color{red}(1.99)       \\ \hline
	6bit     & 68.726\color{red}(2.744) & 73.684\color{red}(1.494) & 75.204\color{red}(2.536)   & 77.016\color{red}(1.085)  & 70.132\color{red}(3.678)     & 77.662\color{red}(1.148)     & 78.086\color{red}(1.306)&64.614\color{red}(4.766)      \\ \hline
	\end{tabular}
	\end{center}
	\end{table*}
	From tabel II we can see that the tha 7bits setting still get good performance for many classic models, the error drop is within $1\%$. However we also can see that the SqueezeNext is more sensitive to bits. it shows that the more light the neural network is, the more sensitive to the schema that symmetric and per-tensor. Also see that the 6bits setting drops more. Maybe need some train. The conclusion is:
	\begin{itemize}
		\item ISQuant can get good accuraccy within $1\%$ drop for many big model.
		\item ISQuant is not very suitable for the light model architecture. 
	\end{itemize}
	The conclusion shows the limitation of our methods, but the feature of free data and fast rate of ISQuant is suitable for some situation. more work should be done.

	The rate of ISQuant is supposed to have the  same time as squant, the table is like table III.
	\begin{table}[htbp]
		\caption{rate of ISQuant}
		\begin{tabular}{|l|l|l|l|l|}
		\hline
		model  & resnet18 & resnet50 & InceptionV3 & SqueezeNext \\ \hline
		Layers & 21   & 54    & 95    & 112         \\ \hline
		ISQuant time(ms)  &84&188 &298&272      \\ \hline
		ZeroQ time(s)    &38&92&136&109       \\ \hline
		GDFQ Time(hour)   &1.7&3.1&5.7&4.8      \\ \hline
		\end{tabular}
	\end{table}
	
	\section{Conclusions}
	A easy deployment schema called ISQuant is proposed in this paper, which replace the quantized convolution's fake quantization with integer computation. The ISQuant can use original convolution computation without additional operation. It achieve good performance in 8bits setting and still have acceptable performance in 7bits setting with less parameters. What's more it is free of data and very fast to do quantization. ISQuant can be applied to many big models with negligible performance drop within $1\%$ at 8bits without the requirement for train data.

	\section*{Acknowledgment}

	\bibliographystyle{unsrt}
    \bibliography{ref}

	\end{document}